\documentclass{article}

\usepackage[
    top=1in,
    bottom=1in,
    left=1in,
    right=1in
]{geometry}

\usepackage{graphicx}
\usepackage{booktabs}
\usepackage{longtable}
\usepackage{array}
\usepackage[numbers,sort&compress]{natbib}
\def\natexlab#1{}
\usepackage{url}
\usepackage{fancyvrb}
\usepackage{amsmath}
\usepackage{float}
\usepackage{wrapfig}
\usepackage{caption}
\usepackage{needspace}
\usepackage{enumitem}
\usepackage{xcolor}
\definecolor{prospector}{HTML}{0B6875}
\definecolor{answergreen}{HTML}{1B6E45}
\definecolor{mutedgrey}{HTML}{6E797D}
\usepackage{fontawesome5}
\newcommand{\codeurl}{https://github.com/lamm-mit/Prospector}

\newcommand{\modelurl}{https://huggingface.co/lamm-mit/Prospector}

\usepackage{fontspec}
\newfontfamily\prospectorfont{texgyreheros}[Extension=.otf, UprightFont=*-regular, BoldFont=*-bold]
\newcommand{\Prospector}{{Prospector}}
\newcommand{\move}[1]{\emph{#1}}   
\definecolor{linkblue}{HTML}{2B4C7E}
\usepackage[colorlinks=true,
            linkcolor=linkblue,
            citecolor=linkblue,
            urlcolor=linkblue]{hyperref}

\makeatletter
\providecommand*{\doi}{}
\renewcommand*{\doi}{\begingroup\hyper@normalise\prospector@doi}
\def\prospector@doi#1{\href{https://doi.org/#1}{doi:\,\nolinkurl{#1}}\endgroup}
\makeatother
\hypersetup{pdftitle={Prospector: learning a compact graph traversal policy for scientific reasoning},
            pdfauthor={Haiqian Yang, Markus J. Buehler}}

\newcommand{\correspondingemail}{mbuehler@mit.edu}
\title{\bf \Prospector{}: learning a compact graph traversal policy for scientific reasoning}
\author{Haiqian Yang, Markus J. Buehler\textsuperscript{*}\\[5pt]
  \normalsize Department of Civil and Environmental Engineering\\
  \normalsize Department of Mechanical Engineering\\
  \normalsize Schwarzman College of Computing\\
  \normalsize Massachusetts Institute of Technology\\
  \normalsize Cambridge, MA, USA\\[5pt]
  \small \textsuperscript{*}Corresponding author: \href{mailto:\correspondingemail}{\texttt{\correspondingemail}}}

\makeatletter
\renewcommand{\@maketitle}{%
  \newpage
  \null
  \vspace*{-0.3in}%
  \begin{center}%
    {\LARGE \@title \par}%
    \vskip 1.2em%
    {\large \lineskip .5em%
      \begin{tabular}[t]{c}\@author\end{tabular}\par}%
  \end{center}%
  \par
  \vskip 1.2em}
\makeatother

\begin{document}

\maketitle

\begingroup
{\centering\small\bfseries ABSTRACT\par}
\medskip
\small
\noindent Scientific reasoning in specialized engineering fields depends on knowledge scattered across the literature, often proprietary and absent from frontier LLM training data. In such domains, frontier models tend to give generic answers. Knowledge graphs (KGs) can ground LLM reasoning. Because a graph is too large to provide in full, typical LLM+KG systems preselect a portion, deciding relevance before reasoning begins. Here we train a compact model, Prospector, to reason over a domain knowledge graph by traversing it step by step, seeking evidence in light of what it has already found. Prospector calls a small set of tools one at a time to progressively build its context, and answers with a chain of evidence that can be checked edge by edge against what its tools retrieved from the graph. On a knowledge graph built from the semiconductor-manufacturing literature, Prospector approaches the strongest off-the-shelf models given the same graph and tools, and it far exceeds off-the-shelf models answering closed-book or from a pre-selected portion of the graph. Although trained only on the semiconductor graph, Prospector transfers to unseen knowledge graphs from three other distinct fields: bio-inspired materials, per- and polyfluoroalkyl substances (PFAS) and materials properties. Served on a single GPU, Prospector answers faster and at a fraction of the cost of a hosted service, while keeping proprietary data on the owner's hardware. We find that the policy's latent state is indicative of its next move on both seen and unseen graphs, and modulating the latent state can steer the policy from one behavior mode to another. These results suggest a practical LLM + KG paradigm for scientific and engineering reasoning over private data, in which a compact model trained on a graph from published literature transfers its traversal to unseen graphs that stay with their owners.
\par\smallskip
\endgroup
\vspace{1mm}

\vspace{-1em}
\begin{figure}[H]
\centering
\captionsetup{font=small}
\includegraphics[width=0.88\textwidth]{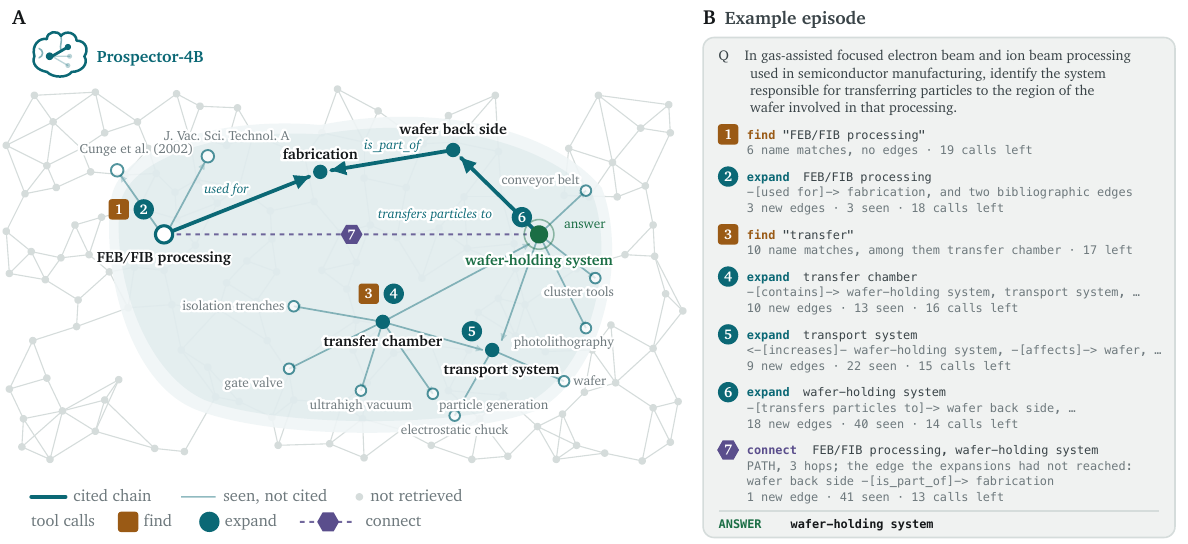}
\caption{\textbf{One episode of \Prospector{} on the semiconductor graph.} \textbf{A}, the region of the graph the policy retrieved, one tool call at a time, with the chain of edges its answer cites. Gray nodes and edges stand schematically for the part of the graph the episode never retrieved. \textbf{B}, the same episode as tool calls and the observations they returned. A sampled episode of \Prospector{}-4B on a held-out question, reproduced in full in Appendix~\ref{app:episodes}.}
\label{fig:overview}
\end{figure}

\section{Introduction}
Automating scientific reasoning has become a concrete engineering goal, and several groups have reported systems that read and synthesize the literature, propose hypotheses, and run experiments and report the results~\citep{ghafarollahi2024sciagents,buehler2024extraction,stewart2026graphagents,skarlinski2024paperqa2,gottweis2025coscientist,lu2024aiscientist}. These systems rest on large language models that internalize knowledge together with reasoning, and ever larger models are trained to serve as both reasoner and encyclopedia~\citep{petroni2019lmkb,roberts2020pack}. This approach is effective while the knowledge is public. On facts that rarely entered the public domain, an off-the-shelf model often answers from general priors instead~\citep{kandpal2023longtail,mallen2023popqa}, so its answer is fluent and plausible but lacks grounding. In semiconductor manufacturing, for example, the process data that would make a model useful stays private to the company that owns it, so it is largely absent from the data off-the-shelf models are trained on~\citep{hsu2026sageqa}.

Augmenting a language model in context is an effective way to ground its reasoning, and one popular method is LLM + knowledge graphs, where the graph supplies the detailed evidence the model lacks~\citep{lewis2020rag,pan2024roadmap}. An answer to an engineering question rests on a chain of relations among materials, mechanisms and process conditions, and these relations are usually scattered across various sources of proprietary documents. A graph stores each relation as an edge with its source, so the chain an answer rests on can be written down and checked against the graph. Because a chain can join edges extracted from different documents, reasoning through the graph can combine evidence spread across sources (Section~\ref{sec:results}).

A knowledge graph is typically too large for a model to read at once, so a graph reasoning system must choose which part to read. The usual approach selects the evidence before reasoning begins: a fixed procedure links the entities the question names, retrieves what lies around them in the graph as edges, passages or community summaries, and places the result in the prompt~\citep{baek2023kaping,he2024gretriever,edge2024graphrag,gutierrez2024hipporag}. The selection is therefore a commitment made before the question has been reasoned about, when least is known about which evidence will matter. When the answer lies several relations away, the entities in between are not necessarily named in the question, which could be easily missed by a selection keyed on the question's entities. The selection is commonly treated as a fix for the context-length problem~\citep{xu2024retrievallong,li2024ragvslong}, but a longer context window enlarges the selection without removing the commitment, and models use long contexts unreliably~\citep{liu2024lostmiddle}. In contrast, avoiding the commitment means retrieving evidence from the graph a step at a time, each step conditioned on what the previous steps returned, and existing systems do so by prompting a frozen model at each step~\citep{sun2024tog,jin2024graphcot}. Nevertheless, the choice of what to read next at each step can instead be trained into a model that is presumably much smaller\footnote{The encyclopedia is the knowledge graph, which is external to the model, so the model needs only the cognitive ability to traverse the graph instead of using its weights to memorize all the facts.} and easier to serve in proprietary situations, and training this choice has been demonstrated on curated benchmark graphs such as Freebase and Wikidata~\citep{jiang2024kgagent,polonuer2026ark,sog2026r1,song2026graphrft,kghopper2026,kgreasoner2026,yan2026eog}. While these methods supply components that a working system can build on, such a system has yet to be demonstrated on real scientific and engineering questions, where it needs to reason over graphs extracted from the literature of different fields and answer with evidence that can be checked. Moreover, where the data are proprietary, it is also preferable for the system to run on the owner's hardware, so that neither the questions nor the graph are sent to a third party.

In this work, we report \Prospector{}, a compact language model post-trained into a traversal policy for scientific knowledge graphs extracted from literature (Section~\ref{sec:recipe}). Traversal is the capability of autonomously choosing what to examine next in light of what the policy has already read. The policy answers by traversing the graph through three tools and asserting a chain of cited edges (Figure~\ref{fig:overview}). 
We find that
\begin{enumerate}[nosep,leftmargin=1.3em,topsep=4pt,partopsep=0pt]
\item \textbf{Selecting a subgraph before reasoning is costly} (Section~\ref{sec:frontier}). Off-the-shelf models answer most questions incorrectly without the graph. They answer better when handed a pre-selected subgraph built around the entity the question is about, but still lose many correct and supported answers they reach with the traversal tools.
\item \textbf{The graph-traversal capability is compact} (Section~\ref{sec:frontier}). Our post-trained 4B model comes close to the strongest frontier model, matches the second and exceeds the other six off-the-shelf models we evaluate, given our graph traversal tools.
\item \textbf{The graph-traversal capability transfers zero-shot to unseen domains} (Section~\ref{sec:transfer}). Our 4B policy trained on the semiconductor graph also approaches the frontier models on three graphs from other fields: bio-inspired materials, per- and polyfluoroalkyl substances (PFAS) and materials properties.
\item \textbf{\Prospector{} is cheap to serve and can run privately} (Section~\ref{sec:cost}). \Prospector{}-4B answers faster and at a much lower cost per question than any of the off-the-shelf models driving the same tools. The same model runs on a laptop without a network connection, so questions and graph data can stay entirely on the owner's machine.
\item \textbf{The latent state predicts the next move, transfers with the policy and can be steered} (Section~\ref{sec:decision_state}). The state from which the policy chooses its next move separates three behaviors: following, backtracking and answering, both on the training graph and on unseen graphs. Modulating the latent state can steer the policy from one behavior mode to another.
\end{enumerate}

\section{Related work}
\label{sec:related}

\subsection{Large language models with knowledge graphs}

Retrieval-augmented generation conditions a language model on evidence retrieved for the query~\citep{lewis2020rag,wu2024stark}. A graph variant retrieves summaries of entity communities in place of passages~\citep{edge2024graphrag}. Systems built on scientific graphs often retrieve before generation begins, so the evidence is selected when least is known about which facts the reasoning will need. SAGE-QA retrieves relation paths from the entities a question names and shares them among prompted agents~\citep{hsu2026sageqa}, and SciAgents places a path sampled between two concepts in the prompt of a multi-agent pipeline~\citep{ghafarollahi2024sciagents}. A second line interleaves retrieval with reasoning, letting a prompted model steer a search over the graph: beam search over relation paths, later extended to documents and to multi-agent retrieval~\citep{sun2024tog,ma2025tog2,chen2025tog3}, iterative reading of the graph through dedicated interfaces~\citep{jiang2023structgpt}, planning with reflection~\citep{chen2024pog}, and decoding constrained to well-formed paths~\citep{li2025dog}. In scientific domains, DeepEvidence and GraphAgents run breadth- and depth-first searches over biomedical and materials graphs~\citep{wang2026deepevidence,stewart2026graphagents}. Graph-CoT is the closest of these to our setting: a prompted model reads a graph one function call at a time, each call conditioned on what the previous calls returned~\citep{jin2024graphcot}. In all of these systems the language model is frozen, and what to read next is set by a fixed search procedure or by a prompt. In contrast, our work trains that choice into the weights of a compact language model (Section~\ref{sec:recipe}).

\subsection{Training large language models on knowledge graphs}
Language models learn to use tools through supervised fine-tuning~\citep{nakano2021webgpt,schick2023toolformer} and reinforcement learning~\citep{guo2025deepseekr1,jin2025searchr1,song2025r1searcher}. Both have been applied to knowledge graphs~\citep{jiang2024kgagent,polonuer2026ark,sog2026r1,song2026graphrft,luo2024rog}. Rewards add terms for a reference path or for exploration~\citep{yan2026eog,sun2026karl}, and earlier work without language models also rewarded plausible answers missing from the graph~\citep{lin2018multihop}. Graph-PRefLexOR trains a model by preference optimization to write a graph and a symbolic abstraction inside its own reasoning~\citep{buehler2025graphpreflexor}. In general-knowledge question answering, models have been trained to query fixed-schema graphs such as Freebase and Wikidata step by step~\citep{sog2026r1,song2026graphrft,kghopper2026,kgreasoner2026}. In this work, the graphs are extracted from scientific literature by a language model~\citep{buehler2024extraction}, with free-text relations and no schema, and Section~\ref{sec:transfer} evaluates the trained policy zero-shot on three graphs it was not trained on.

\subsection{Verifying answers against evidence}
Whether a generated claim is supported by its sources is usually judged by human raters~\citep{rashkin2023ais} or by language models~\citep{gao2023alce,gao2023rarr,min2023factscore}, and model judges carry biases of their own, such as favoring longer answers~\citep{zheng2023judge}. On knowledge graphs, KGARevion has a fine-tuned model review each generated triple against the graph~\citep{su2025kgarevion}, and SAGE-QA, built on the semiconductor graph used here, was scored by LLM judges~\citep{hsu2026sageqa}. Graph structure has also been used to guide which hypotheses are assembled~\citep{stewart2026hypergraph}. In this work, support is a set-membership check: every cited edge must have been returned by the tools, and the answer must appear in the cited edges (Figure~\ref{fig:episode}B, Section~\ref{sec:gate}). The same check is used in the training reward.

\section{Results}
\label{sec:results}
We study question answering over a knowledge graph that a language model extracted from semiconductor-manufacturing literature. \Prospector{} answers a question by traversing the graph one tool call at a time, so that each step is conditioned on what the previous calls returned. One of the held-out questions, for example, asks: \emph{when center gas flow is adjusted to correct within-die pattern dimension variations, those adjustments can produce process by-products that influence the dimensions; which processing stage is therefore directly implicated as containing those by-products?} The question names \texttt{center gas flow}, and answering it requires three edges:

\begin{quote}
\ttfamily\small
center gas flow --[tuned for]--> CD uniformities\\
etch by-products --[affects]--> CD uniformities\\
HAR processing --[contains]--> etch by-products
\end{quote}

\noindent where each edge of the graph is a directed triple of two entity names and a free-text relation.\footnote{As opposed to previous literature typically in LLM + KG, our graphs are represented in free text form and have no schema, as they are extracted directly from scientific literature: the extractor wrote each relation name and type label while reading the passage, so relation names are ordinary English fragments such as \texttt{affects}, \texttt{used in} and \texttt{includes}.} Read as a chain, the edges state that center gas flow is tuned for critical-dimension (CD) uniformity, that etch by-products affect that uniformity, and that high-aspect-ratio processing (HAR) contains those by-products. The answer is therefore \texttt{HAR processing}. The three edges were extracted from three different documents, a textbook on atomic layer processing~\citep{lill2021atomiclayer}, a textbook on dry etching~\citep{nojiri2015dryetching} and a review of plasma etching~\citep{oehrlein2024plasmaetching}. Asked the same question without the graph, all eight off-the-shelf models we evaluate named a generic etching stage.



\subsection{A compact policy approaches the strongest frontier models}
\label{sec:frontier}

\begin{figure}[h!]
\centering
\includegraphics[width=\textwidth]{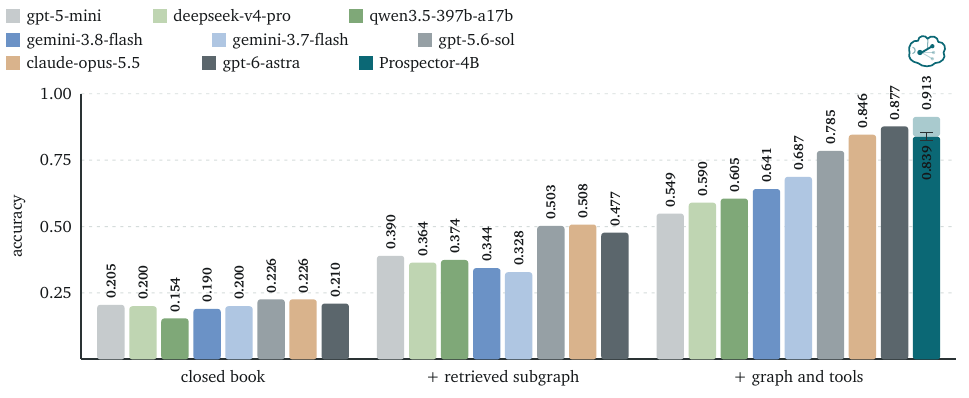}
\caption{\textbf{Tool-driven traversal versus fixed retrieval.} Success on 195 held-out items. Each off-the-shelf model appears three times: answering closed book, handed a retrieved subgraph fixed before generation around the entity regarding the question, and driving the tools itself. Every model that drives tools uses the same three tools (\texttt{find, expand, connect}). Models are ordered by their accuracy with the tools. The lighter segment on top of \Prospector{}-4B's bar is its accuracy with the trained verifier (Section~\ref{sec:decision}).}
\label{fig:frontier}
\end{figure}

We compare \Prospector{}-4B with eight off-the-shelf models: \texttt{gpt-6-astra}, \texttt{gpt-5.6-sol}, \texttt{gemini-3.7-\allowbreak flash}, \texttt{gemini-3.8-flash}, \texttt{claude-opus-5.5}, \texttt{deepseek-v4-pro}, \texttt{qwen3.5-397b-a17b} and the smaller \texttt{gpt-5-mini}, on the same 195 held-out items (Figure~\ref{fig:frontier}). Each off-the-shelf model is evaluated under three conditions: closed book, with a subgraph retrieved and given to the model in the prompt before generation, and driving the same three tools as the \Prospector{} policy (Sections~\ref{sec:baselines} and~\ref{sec:harness}).

\Prospector{}-4B, trained with the recipe of Section~\ref{sec:recipe}, reaches an accuracy of $0.839 \pm 0.015$ in a single greedy pass, against 0.877 for \texttt{gpt-6-astra}, the strongest frontier model, 0.846 for \texttt{claude-opus-5.5}, and 0.549 to 0.785 for the other six off-the-shelf models. 
In Section~\ref{sec:decision} we show that sampling and verifier selection raise the accuracy further, to 0.913.

Without the graph, the eight off-the-shelf models score alike: their answer accuracy ranges from 0.154 to 0.226. The differences in general capability that usually separate these models do not appear on a highly focused domain corpus. A subgraph retrieved before generation helps far less than driving the tools, even when the retrieval is handed the entity the question is about (Section~\ref{sec:retrieved}). Retrieval raises \texttt{gpt-5.6-sol} to 0.503 against 0.785 with the tools, and \texttt{gpt-6-astra} to 0.477 against 0.877. Across the eight off-the-shelf models the tools add between 0.16 and 0.40 over retrieval, a difference indicating the cost of selecting evidence before the question has been reasoned about.

\subsection{How the trained policy traverses the graph call by call}
\label{sec:harness}
\begin{figure}[!htb]
\centering
\includegraphics[width=\textwidth]{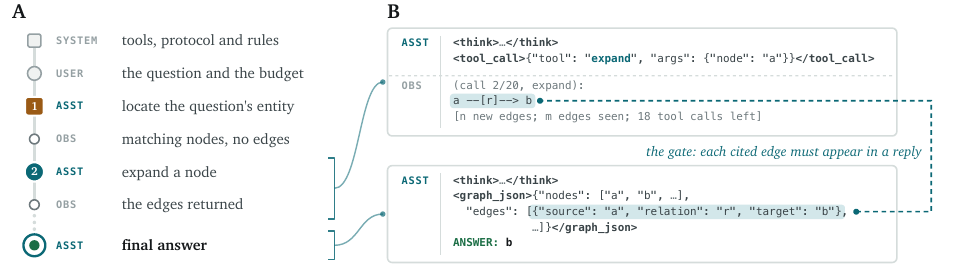}\\[4pt]
\includegraphics[width=\textwidth]{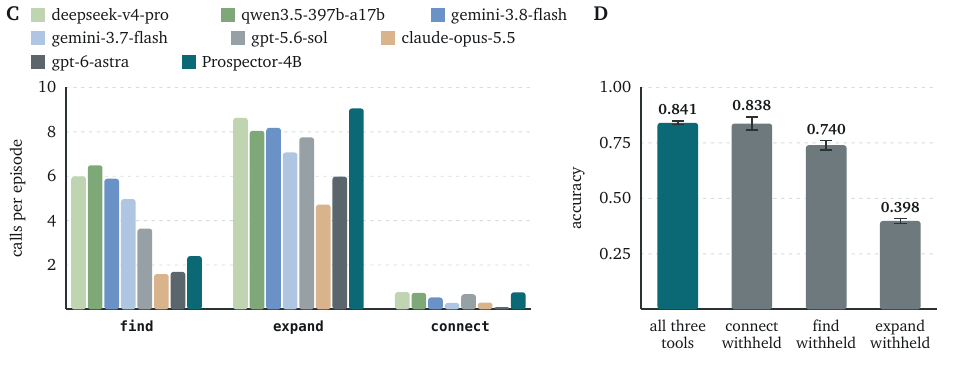}
\caption{\textbf{The messages of an episode, tool calls per episode and the effect of withholding each tool.} \textbf{A}, one episode from the system prompt to the final answer, in which a call and its reply repeat up to the budget of 20 calls. \textbf{B}, the two bracketed groups of A enlarged. \texttt{a}, \texttt{r} and \texttt{b} stand for node and relation names, and the dashed line joins a cited edge to the reply that returned it. The system prompt is reproduced in Appendix~\ref{app:prompt}. \textbf{C}, calls per episode to each of the three tools. \textbf{D}, accuracy of \Prospector{}-4B when one tool is withheld at evaluation. A withheld tool is removed from the tool list the policy is shown and a call to the withheld tool is refused by the harness. Whiskers are one standard deviation over three runs.}
\label{fig:episode}
\label{fig:tools}
\end{figure}

The \Prospector{} policy reaches the graph only through three tools, and the graph itself is never fully placed in its context (Figure~\ref{fig:episode}A,B). \texttt{find} locates nodes by name and returns no edges, \texttt{expand} returns the edges of a located node in both directions, and \texttt{connect} returns up to three shortest paths between two located nodes. On each turn the policy writes a short reasoning block and then either calls one tool or gives the final answer, and the result of a call arrives as the next message and is appended in the model's context. A \textit{seen set} is formed by all the edges returned by a full episode's tool calls\footnote{In this paper, an episode is capped at 20 calls and 400 seen edges.}. Sections~\ref{sec:tools} and~\ref{sec:episode} specify the tools and the episode loop, and Appendix~\ref{app:episodes} reproduces a complete episode.

When we analyze the frequency of each type of tool, we find that the trained policy uses the \texttt{expand} tool more often than any other model (Figure~\ref{fig:tools}C). On average, \Prospector{}-4B expands 9.1 times per episode, against 4.7 to 8.6 for the off-the-shelf models, and calls \texttt{find} 2.4 times, within their range of 1.6 to 6.5. It calls \texttt{connect} 0.8 times per episode.

Withholding each tool in turn, against a separate set of three runs with all tools available, shows that the policy depends on \texttt{expand} and hardly on \texttt{connect} (Figure~\ref{fig:tools}D). Withholding \texttt{connect} leaves the accuracy almost unchanged, 0.838 against 0.841, and in fact, the policy stops calling \texttt{connect} entirely. Removing \texttt{find} costs 0.101: the policy still attempts to call \texttt{find} in 0.37 of episodes, while those calls are refused by the harness, which sends a prompt back with the list of the remaining tools, and often the policy then substitutes \texttt{find} with \texttt{expand}, which rises from 8.9 to 11.1 calls per episode, retrieving the answer in 0.83 of episodes instead of 0.92. Removing \texttt{expand} reduces gated success to 0.398, which is intuitive since \texttt{expand} is the only tool that enumerates a node's neighborhood: \texttt{find} returns no edges, and \texttt{connect} requires two nodes that the policy needs to locate before calling it. When \texttt{expand} is removed from the allowed tool list, the policy substitutes as well, calling \texttt{find} 7.8 and \texttt{connect} 6.1 times per episode against 2.5 and 0.7 with all three available, and retrieves the answer in 0.39 of episodes against 0.92.

\subsection{Choosing the answer among retrieved candidates}
\label{sec:decision}

\begin{figure}[!t]
\centering
\includegraphics[width=\textwidth]{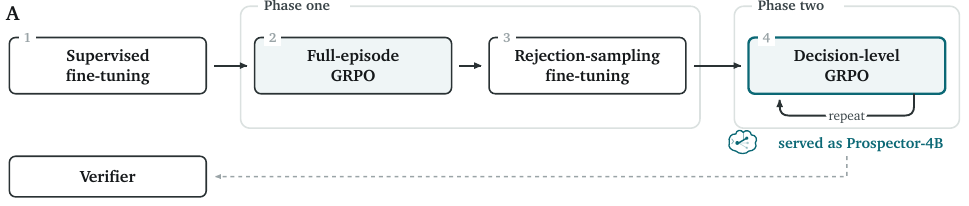}\\[4pt]
\includegraphics[width=\textwidth]{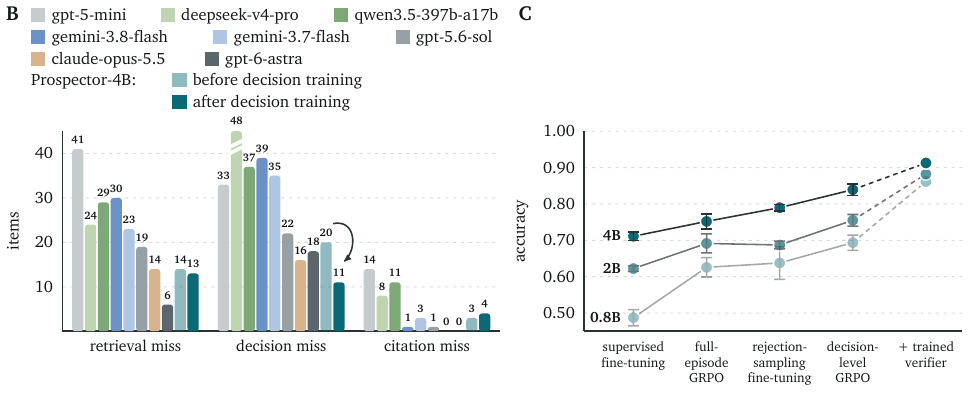}
\caption{\textbf{The training recipe, misses by cause and their recovery by training and by selection.} \textbf{A}, the stages of the training recipe in order (Section~\ref{sec:recipe}). \textbf{B}, the misses of the eight off-the-shelf models of Figure~\ref{fig:frontier}, and of the \Prospector{}-4B policy before and after the decision-level rounds, split by cause. \textbf{C}, accuracy of a greedy pass after each stage of A, for the same recipe at three model sizes. The vertical axis starts at 0.45; whiskers are one standard deviation.}
\label{fig:recipe}
\label{fig:selection}
\end{figure}

In the final answer turn of an episode, choosing among all the candidates in its context is a bottleneck shared by every model evaluated. As an episode ends with an answer turn, the policy asserts one entity and cites the chain of edges as supporting evidence (Figure~\ref{fig:episode}B), while the candidates are all the entities the episode has seen by then. When the final answer is wrong, we classify each failure by the stage that failed (Figure~\ref{fig:selection}B). A \emph{retrieval miss} never retrieved a correct answer: no episode tool call returned any edge that ever touches the correct answer. A \emph{decision miss} retrieved the correct answer as a candidate but asserted a different entity in the final answer turn. A \emph{citation miss} hallucinated evidence that does not support the answer, either an edge the episode never saw or a chain that does not contain the entity asserted.\footnote{The citation test is applied first, so the other two classes hold only misses whose citations the gate accepted.}

Decision misses are the largest class for the 4B policy before the decision-level rounds, 20 of its 37, and the largest for seven of the eight off-the-shelf models; \texttt{gpt-5-mini} misses more often in retrieval, 41 items against 33. Across the off-the-shelf models, retrieval misses fall from 41 of \texttt{gpt-5-mini}'s 88 to 6 of \texttt{gpt-6-astra}'s 24, while decision misses stay between 16 and 48. To address the decision failures, we employ a second reinforcement-learning phase that trains the policy from the states where an episode's outcome is decided (Section~\ref{sec:rl}), as shown in the second phase of the recipe in Figure~\ref{fig:recipe}A.
Here the prompts are the policy's own episodes, cut at the turn where the answer is decided: from each cut state sixteen continuations are sampled and run to the end and only the continuation carries the loss.
Because the states are episodes the policy wrote, training on them leaves them stale, and the round is repeated on states rebuilt from the policy it has just produced.
The decision-level rounds reduced decision misses from 20 to 11 items (Figure~\ref{fig:selection}B), and raised the single greedy pass accuracy from $0.790 \pm 0.009$ after the first phase (stage 3) to $0.839 \pm 0.015$ after the decision-level rounds, against $0.711 \pm 0.012$ after supervised fine-tuning on the demonstrations (Figure~\ref{fig:selection}C) and $0.039 \pm 0.008$ for the base model driving the same tools.

Nevertheless, when we sample a pool of 32 episodes per item, within the pool the system holds a correct and grounded answer for 192 of the 195 items, which is much better than a single greedy sampling. Therefore, we further train a verifier (Section~\ref{sec:verifier}) to choose the correct candidate from the pool, raising the accuracy further to 0.913, against 0.839 for the single pass and 0.856 for a simple majority vote over the same pool (Figure~\ref{fig:selection}C).

The same recipe applied to post-training \texttt{Qwen3.5-0.8B} and \texttt{Qwen3.5-2B} raises them from 0.487 and 0.622 after supervised fine-tuning to 0.694 and 0.755 after the decision-level rounds, and each size's own verifier raises them further, to 0.862 and 0.882 (Figure~\ref{fig:selection}C).

\subsection{The graph-traversal capability transfers zero-shot to unseen domains}
\label{sec:transfer}
\begin{table}[!t]
\centering
\captionsetup{font=small}
\caption{\textbf{Zero-shot transfer to three unseen graphs.} Accuracy on the bio-inspired-materials, PFAS and materials-properties graphs, at the 20-call budget; $\pm$ spans three runs.}
\label{tab:transfer}
\footnotesize
\setlength{\tabcolsep}{2.5pt}
\begin{tabular}{l ccc ccc ccc}
\toprule
 & \multicolumn{3}{c}{Bio-inspired materials} & \multicolumn{3}{c}{PFAS} & \multicolumn{3}{c}{Materials properties} \\
\cmidrule(lr){2-4} \cmidrule(lr){5-7} \cmidrule(lr){8-10}
 & Tools & Retrieval & \shortstack[c]{Closed\\book} & Tools & Retrieval & \shortstack[c]{Closed\\book} & Tools & Retrieval & \shortstack[c]{Closed\\book} \\
\midrule
\Prospector{}-4B & $0.803 \pm 0.017$ & --- & --- & $0.766 \pm 0.022$ & --- & --- & $0.809 \pm 0.011$ & --- & --- \\
\texttt{gpt-6-astra} & 0.857 & 0.681 & 0.176 & 0.846 & 0.294 & 0.206 & 0.865 & 0.435 & 0.148 \\
\texttt{gemini-3.7-flash} & 0.762 & 0.519 & 0.176 & 0.719 & 0.254 & 0.197 & 0.771 & 0.336 & 0.143 \\
\texttt{gemini-3.8-flash} & 0.681 & 0.543 & 0.152 & 0.662 & 0.254 & 0.197 & 0.700 & 0.372 & 0.161 \\
\texttt{gpt-5.6-sol} & 0.757 & 0.700 & 0.181 & 0.759 & 0.329 & 0.189 & 0.776 & 0.475 & 0.148 \\
\bottomrule
\end{tabular}
\end{table}
\begin{figure}[!t]
\centering
\includegraphics[width=\textwidth]{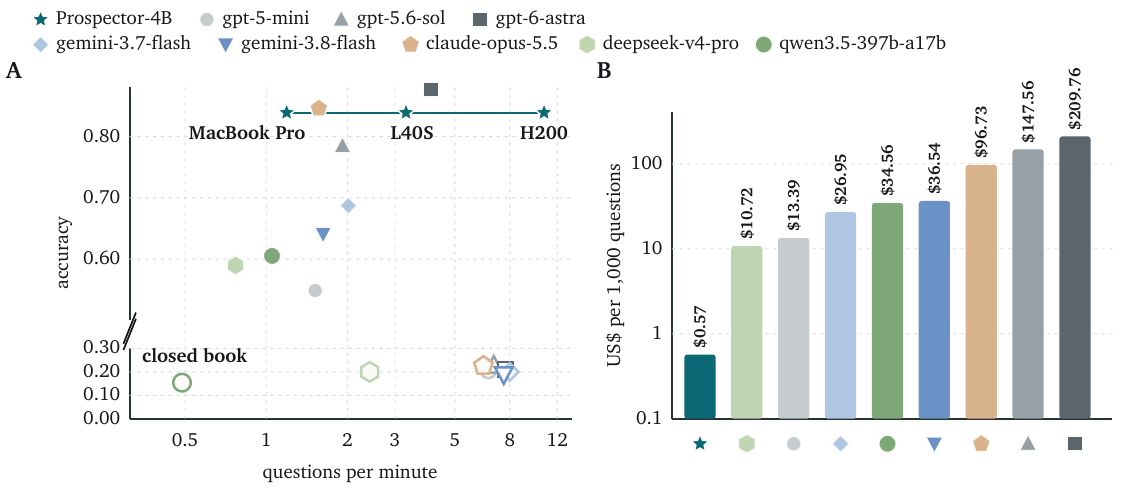}
\caption{\textbf{Cost of one question in time (A) and in money (B).} \textbf{A}, accuracy against questions answered per minute with one episode at a time, on a log scale. The vertical axis is broken between 0.30 and 0.50. Hollow markers indicate the closed-book performance. The line joins \Prospector{}-4B served on three machines. \textbf{B}, US dollars per thousand questions for one pass on a log scale, the bars rising from \$0.10, priced from a rented GH200 at \$2.29 per hour for \Prospector{}-4B with the benchmark's episodes run concurrently. For the off-the-shelf models, prices per million input and output tokens are the providers' posted rates.}
\label{fig:cost}
\end{figure}

To test whether the policy trained on one scientific knowledge graph can generalize to graphs in other domains that it was not trained on, we apply this policy to three separate graphs covering bio-inspired materials, PFAS and materials properties, which were extracted from other literature. 
With its weights frozen, \Prospector{}-4B zero-shot reaches an accuracy of $0.803 \pm 0.017$, $0.766 \pm 0.022$ and $0.809 \pm 0.011$ on the three graphs (Table~\ref{tab:transfer}), second on each to \texttt{gpt-6-astra} at 0.857, 0.846 and 0.865, and ahead of or level with the other three off-the-shelf models: \texttt{gemini-3.7-flash} at 0.762, 0.719 and 0.771, \texttt{gpt-5.6-sol} at 0.757, 0.759 and 0.776, and \texttt{gemini-3.8-flash} at 0.681, 0.662 and 0.700. On these additional graphs the off-the-shelf models score closed-book between 0.143 and 0.206, and a subgraph selected in advance around the question's entity costs them 0.06 to 0.24 in accuracy on the bio-inspired-materials graph against driving the tools, and 0.30 to 0.55 on the PFAS and materials-properties graphs.

\subsection{\Prospector{} is cheap to serve and can run privately}
\label{sec:cost}

\Prospector{}-4B is faster and less expensive to serve than any of the eight off-the-shelf models driving the same tools (Figure~\ref{fig:cost}). 
We benchmark the speed with one episode at a time with no parallelization. 
On an H200, the policy answers 10.7 questions per minute, against 4.1 for \texttt{gpt-6-astra} and 0.8 to 2.0 for the other seven; \Prospector{}-4B reaches 3.3 on an L40S, and 1.2 on a MacBook M4 Pro, which runs without a network connection and at a negligible cost per question. Priced on a rented GH200, one pass costs \$0.57 per thousand questions with episodes run concurrently, against \$10.72 for \texttt{deepseek-v4-pro}, \$13.39 for \texttt{gpt-5-mini}, \$26.95 for \texttt{gemini-3.7-flash}, \$34.56 for \texttt{qwen3.5-397b-a17b}, \$36.54 for \texttt{gemini-3.8-flash}, \$96.73 for \texttt{claude-opus-5.5}, \$147.56 for \texttt{gpt-5.6-sol} and \$209.76 for \texttt{gpt-6-astra} through their APIs.

\subsection{Reading and steering the next move from the policy's latent state}
\label{sec:decision_state}
\begin{figure}[!t]
\centering
\includegraphics[width=\textwidth]{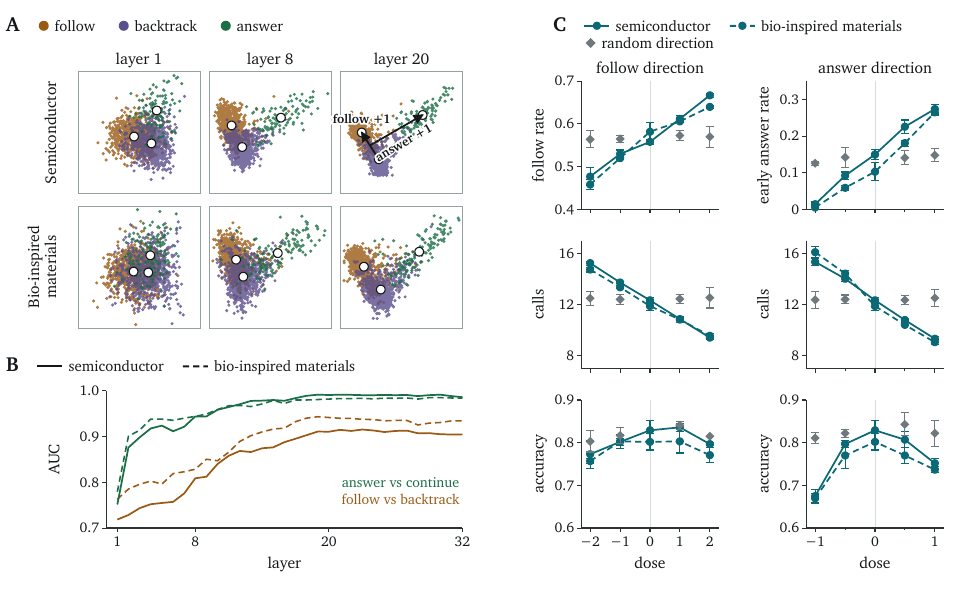}
\caption{\textbf{The next move can be predicted and steered from \Prospector{}-4B's latent state.} \textbf{A}, latent states at the start of a turn, by the move taken next, in the plane that separates the three moves on the semiconductor graph, at three layers; the states of the bio-inspired-materials graph are projected onto the same plane. Arrows mark one dose unit of each direction. \textbf{B}, AUC by layer of a linear reading of each decision fitted on the semiconductor graph, held out on that graph (solid) and applied to the bio-inspired-materials graph (dashed). \textbf{C}, the share of the next moves that follow the observation of the last turn, or the share of episodes answered prematurely, tool calls per episode and accuracy against the dose added along the follow direction (left) and the answer direction (right), on the semiconductor (solid) and bio-inspired-materials (dashed) graphs. Gray diamonds indicate a baseline where the policy is steered in a random direction of the same norm ($n=3$). Points are means over three runs; whiskers are one standard deviation.}
\label{fig:decision_state}
\end{figure}

Next, we ask whether the model has formed a generalizable internal representation of graph traversal across the trained and the unseen graphs. To do so, we analyze the latent state of \Prospector{}-4B at the end of an observation, right before the model outputs the first token of the next turn (Section~\ref{sec:decision_state_methods}). Notably, the internal representation segregates into three distinct clusters: \move{follow}, which expands a node on the observation of the last turn, \move{backtrack}, which expands a node from an earlier turn or from the question, and \move{answer}, which gives the final answer (Figure~\ref{fig:decision_state}A). The three moves separate with network depth: at layer 1 the classes overlap, by layer 8 answering has moved apart, and by layer 20 following and backtracking have separated as well, on the unseen graph as on the training graph. A linear reading of each decision, fitted on the semiconductor graph, predicts the model's next move on the bio-inspired-materials graph without refitting: at layer 20 it separates answering from continuing with an AUC of 0.99 on both graphs, and following from backtracking with 0.91 on the semiconductor graph and 0.94 on the bio-inspired-materials graph (Figure~\ref{fig:decision_state}B). These results indicate that the model's intention for the next move can be read out before the move is made.

Given such a distinct internal representation, we ask whether we can intentionally bias the policy from one behavior mode to the other by modulating its latent state. For example, the trained model follows its chain of evidence at a rate of around 0.55 and backtracks often, and we ask whether it can be made to follow its chain of evidence more confidently and backtrack less. This is achieved by adding to the latent state a vector pointing from the \move{backtrack} center to the \move{follow} center, which also lowers the mean number of tool calls per episode without a loss of accuracy at a moderate dose. This applies both to the semiconductor graph the policy was trained on and to the unseen bio-inspired-materials graph (Figure~\ref{fig:decision_state}C); the same trend can also be seen on the PFAS and materials-properties graphs.
Furthermore, the same operation along the direction of the \move{answer} cluster advances or delays the final answer (Figure~\ref{fig:decision_state}C). Both operations work on the training graph and on entirely unseen graphs, which indicates that the policy's internal representation of evidence seeking is general, and that it can be modulated to change its traversal behavior.

\section{Discussion}

\paragraph{Separation of knowledge from reasoning.}
Scientific work draws on two capabilities: holding a literature, and knowing how to move through it. Expertise rests mostly on the second. Being a scientist is not about the memorization of textbooks but a practiced way of working, which is why a researcher can enter an unfamiliar field and contribute to it; few experts can recite a paper, and all of them know where to look. A language model keeps both capabilities in one set of weights, and buys the second by scaling the first. The results here suggest the two can be held apart. Closed book, models of very different size answer these questions alike and poorly, and it is only once they must search the graph that they separate (Figure~\ref{fig:frontier}); our fine-tuned graph-traversal policy is small enough to serve on a laptop, and it carries its search to graphs it has never read (Section~\ref{sec:transfer}). The latent state is indicative of the policy's next move, both on the training graph and on a graph it has never read, and modulating the latent state adjusts the traversal behavior mode (Section~\ref{sec:decision_state}). 
Holding the knowledge and the reasoning apart also means the external knowledge graph can be corrected or extended without retraining the model.

\paragraph{Serving on private data.}
The transfer results demonstrate that this system can be useful especially when a technical corpus cannot be shared. The policy runs frozen on graphs it was never trained on and reaches reasonable accuracy (Section~\ref{sec:transfer}), so a private graph can be served by a model that has never read it. Serving then also stays local: \Prospector{}-4B can answer on a laptop without a network connection (Section~\ref{sec:cost}), so no question, edge or answer leaves the owner's machine.

\paragraph{Extension to hypothesis generation.}
The questions studied here carry an answer set computed from the graph, whereas an open-ended question does not carry such an answer set. Groundedness remains available, since the gate verifies only that the cited edges were retrieved and that verification requires no answer key. 
An extension to hypothesis generation needs a second criterion: whether a claim deserves an experiment. A language model scoring against a set of explicit rubrics is one possible future direction, on criteria such as mechanistic plausibility, novelty against the literature and experimental cost.

\section{Conclusion}
The knowledge that is critical in a specialized domain is often proprietary and remains with its owner, whether a scientist, a research group or a company. In such domains, the off-the-shelf LLMs tend to provide generic answers that are not useful for technical experts; the useful knowledge lives instead in a large number of technical documents that the owners prefer not to share publicly.
In this work, we build \Prospector{}, an LLM policy post-trained to answer scientific questions grounded in the technical corpora.
\Prospector{} is a compact policy that traverses through tool calling, on free-text knowledge graphs extracted from scientific literature, prospecting new evidence turn by turn, given what it already found; it approaches the best of frontier models, and much faster and cheaper, at a size that runs on a laptop. 

To train the \Prospector{} model, we develop a training method consisting of different phases, tackling different failure modes of graph traversal, such as a decision-level reinforcement-learning phase that improves the model's ability to pick the correct, grounded final answer given the full evidence it retrieved in its context.
Trained only on a semiconductor manufacturing graph, with frozen weights the system also approaches frontier performance zero-shot on graphs from entirely unseen fields, such as bio-inspired materials, PFAS, and materials properties, showing a general, transferable graph-traversal capability learned through our post-training.

This transferable graph-traversal capability can be further visualized by probing the policy's latent state, which segregates into distinct clusters indicating the different traversal modes, from which its next move could be predicted before the move even takes place; this internal representation of graph traversal carries over to unseen graphs, and can be used to steer the policy from one behavior mode to another.
The \Prospector{} system sheds light on a new paradigm of machine intelligence in autonomous, proprietary scientific reasoning. 

\section{Materials and methods}
\label{sec:methods}

\subsection{Graph datasets}
\label{sec:kgdata}
Four graphs are used in this study.
The semiconductor graph, used for all training and for the main evaluation, was built for SAGE-QA~\citep{hsu2026sageqa} by a generative extraction pipeline~\citep{buehler2024extraction,buehler2024cephalo}. 
\texttt{bio-graph-1K} is extracted from roughly a thousand bio-inspired-materials papers by the same generative extraction pipeline~\citep{buehler2024extraction} and released with SciAgents~\citep{ghafarollahi2024sciagents}. 
The PFAS and materials-properties graphs were released with GraphAgents~\citep{stewart2026graphagents} and built by the same pipeline.

\subsection{Benchmarks}
\label{sec:questions}
To generate the question set, a gold path is first sampled as a self-avoiding walk over domain nodes, excluding hubs and bibliographic nodes. 
A teacher model then phrases a question that names the start entity and no intermediate node, in the form of a domain exam question, and questions naming any intermediate node are filtered out.
The answer set is the set of endpoints reached by the gold relation sequence over the full graph.
Items naming the gold answer in the question are filtered out.
The procedure yields 195 held-out items on the semiconductor graph, 210 on \texttt{bio-graph-1K}, 228 on the PFAS graph and 223 on the materials-properties graph.
Separately, 3,971 training items are generated, and no training item shares a gold path with any evaluation item. 

\subsection{Agent harness}
\subsubsection{Tools}
\label{sec:tools}
The policy interacts with the graph only through tool calls, and the full graph itself is never placed in the context window. 
Three tools are provided, and a tool withheld for an ablation is removed from the list the policy is shown and refused on the harness side if it is ever called by the policy. 
\texttt{find} receives a name and returns the matching node names, each with its type label and degree. 
\texttt{expand} receives the name of a node the policy has already located and returns the edges linked to that node.
\texttt{connect} receives two located nodes and returns up to three shortest paths of at most three hops between them.

\subsubsection{Episode}
\label{sec:episode}
At the beginning of each episode, the policy locates the entity named in the question with \texttt{find} and expands outward one tool calling turn at a time.
An episode interleaves reasoning and action in the manner of ReAct~\citep{yao2023react}. On each turn the policy may write a short reasoning block in \texttt{<think>} tags, and then either call exactly one tool, in \texttt{<tool\_call>} tags, or give its final answer.
A tool call is always answered by an observation from the harness (the results returned by the tool), which arrives as the next message and begins the following turn.
The final answer is a single message carrying the reasoning block, the cited chain in \texttt{<graph\_json>} tags, which is a chain of relations from the start entity to the answer node, and the asserted answer on a line beginning \texttt{ANSWER:}.
Figure~\ref{fig:episode}A,B shows the messages of an episode and the blocks of its turns.

\subsubsection{Gate}
\label{sec:gate}
The accuracy in this paper is measured as a gated success rate, where the gate scores an answer by set membership: an answer passes the gate only if every edge it cites lies in the episode's seen set and the answer lies within the cited edges. An answer is correct when it contains a member of the item's answer set.

\subsection{Baselines}
\label{sec:baselines}
\subsubsection{Closed book}
\label{sec:closedbook}
The closed-book baseline receives the question alone, with no additional information from the graph or tool calling. This condition measures how much of a graph's content the model already holds.

\subsubsection{Retrieved subgraph}
\label{sec:retrieved}
The retrieved-subgraph baseline receives a subgraph in the prompt and answers in a single pass.
Two selection rules are used, and each graph is given the one whose retrieved edges contain the gold chain more often.
Both rules are directly seeded with the entity the question is about, while \Prospector{} would need to first locate that entity itself through \texttt{find}. $k$-hop orders edges by hop distance from that entity; Personalized PageRank runs a walk that follows an edge with probability 0.85 and otherwise returns to that entity, then scores each edge by the mass on its two endpoints. 

\subsection{Training recipe}
\label{sec:recipe}
Throughout, low-rank adapters are employed to fine-tune the models~\citep{hu2022lora}: the base model's weights are held fixed and a rank-16 update is trained for each of its attention and feed-forward projections.
The policy is trained in the stages of Figure~\ref{fig:recipe}A. First, supervised fine-tuning on a frontier model's demonstrations gives the policy the format and the tool grammar, and then two phases of reinforcement learning follow, the first phase trains the full episode and the second phase trains the decision.
In the first phase the round is followed by rejection-sampling fine-tuning, a supervised stage that trains a fresh adapter from the base model on the successful episodes the round produced, so the round's contribution reaches the second phase as data rather than as weights. 
Because the second phase's prompts are episode prefixes the policy itself produced and a changed policy produces different ones, it is useful to re-roll the episodes and repeat this phase.
The served model is the first phase's supervised adapter with one adapter for each decision-level round applied, and a final stage trains the verifier that selects among sampled episodes. Sections~\ref{sec:sft} to~\ref{sec:verifier} describe the stages, and Section~\ref{sec:config} lists the evaluation settings.

\subsubsection{Supervised fine-tuning}
\label{sec:sft}
Demonstration episodes are generated by \texttt{gemini-3.7-flash}, driving the same tools through the same harness on training questions.
Each training question contributes at most one demonstration: the teacher attempts every question once, and a second attempt is drawn for the questions its first attempt failed.
The initial supervised fine-tuning trains the policy on demonstration episodes with a next-token loss. Each turn consists of the reasoning text and the tool call that follows the reasoning, and the tool replies are masked out of the loss.
In between the reinforcement learning phases, a rejection-sampling round is applied~\citep{yuan2023rft}, which trains from a base checkpoint on the episodes sampled from the policy~\citep{zelikman2022star,singh2024beyond,grattafiori2024llama3,guo2025deepseekr1}. At most two episodes per training question are kept, the shortest first, and they are blended with the teacher demonstrations.

\subsubsection{Reinforcement learning}
\label{sec:rl}
In the reinforcement-learning stage the policy generates its own episodes and is updated with group-relative policy optimization (GRPO)~\citep{shao2024deepseekmath,guo2025deepseekr1}. Two phases are employed here. They differ in where the group is formed and in which turns carry the loss, and share the rest of the setup. Sixteen rollouts are sampled per group from the current policy and run to the end of the episode, under the same loop, tools and budget as evaluation. Each is scored by the gate: a gated success receives a reward of one, and any other rollout receives a quarter of its gold-path recall, the fraction of the gold path's edges that its calls returned. A divergence penalty is used to keep the policy near the checkpoint the round starts from. Before each round, the policy is first sampled over the round's candidates, and only those it solves sometimes but not always are kept for training.
The full-episode training starts from the question and every turn carries the loss, whereas the second phase trains only the decision: the group is formed at a decision state and only the continuation carries the loss, as in earlier work that restarts from intermediate states~\citep{florensa2017reverse,resnick2018backplay,salimans2018montezuma,xi2024r3}.
To build the data for the decision phase, four passes are rolled over the training questions, their outcomes are labeled by the classes of Section~\ref{sec:decision}, and each episode is cut at the turn where its outcome was decided, in one of two ways.
An \emph{answer state} is cut before the final turn of an episode that had already retrieved a correct answer; an \emph{expansion state} is cut in an episode that never retrieved the answer, before the turn that followed its first sight of a node one hop from it.

\subsubsection{Verifier}
\label{sec:verifier}
Sampling a pool of episodes for one question yields a correct and grounded answer far more often than a single pass (Section~\ref{sec:decision}), so a final stage trains a model to choose among them~\citep{cobbe2021verifiers}. The verifier is a separate yes/no classifier fine-tuned from the policy's base model. Each candidate is reduced to the question, the answer the episode asserted and the edges it cited. The verifier is asked whether the answer is correct and supported by the cited edges as a connected chain from the question's subject to the answer, and its score is the margin $\log P(\text{yes}) - \log P(\text{no})$ on the reply token, read with two log-probability calls.

\subsection{Reading and steering the latent state}
\label{sec:decision_state_methods}
We record the latent state after each decoder layer at the token that opens a turn, before the policy writes anything of the next turn. We keep the turns that follow an \texttt{expand} reply and label each by the move it makes: a \move{follow} expands a node named on that page, a \move{backtrack} expands a node named on an earlier page or in the question, and an \move{answer} ends the episode. The semiconductor graph contributes 705, 668 and 104 turns of the three classes, and the bio-inspired-materials graph 696, 741 and 135.

Panel A of Figure~\ref{fig:decision_state} is the plane of the two Fisher discriminant axes of the three classes, fitted on the semiconductor states; the bio-inspired-materials states are zero-shot projected onto the same plane. Panel B scores two one-dimensional readings fitted the same way at every layer, \move{follow} against \move{backtrack} and \move{answer} against \move{continue} (either \move{follow} or \move{backtrack}), by the area under the ROC curve: held out on the semiconductor graph, and zero-shot on the bio-inspired-materials graph.

The two directions of panel C of Figure~\ref{fig:decision_state} are unit vectors in the raw latent space at layer 20 of the semiconductor states: the \move{follow} direction points from the \move{backtrack} mean to the \move{follow} mean, and the \move{answer} direction from the mean of the other two moves to the \move{answer} mean. A dose adds the unit vector to the output of layer 20, starting from the header token and at every generated token afterwards. The random baseline is a unit vector drawn from a standard normal distribution, added at each direction's dose unit.

\subsection{Configuration}
\label{sec:config}
Every evaluation uses a budget of 20 tool calls, a cap of 400 seen edges, a repeat cap of three on identical calls, 40 edges per page and a hub threshold of 150. \Prospector{} is served with a 16,384-token context window, the length it was trained at, except on the PFAS graph, where hub summaries are long enough that a 32,768-token window is used (Table~\ref{tab:transfer}). Each turn of \Prospector{} is generated with at most 1,024 new tokens, decoded greedily. \Prospector{}-4B is stage 4 of Figure~\ref{fig:recipe}A: the rank-16 adapter of stage 3 on \texttt{Qwen3.5-4B}, with the adapters of the two decision-level rounds applied.
The off-the-shelf models run at their providers' default reasoning effort.

\section*{Use of Generative AI}
\vspace{-0.7\baselineskip}
The authors used Claude Code (Anthropic; Claude Opus 5, Opus 5.5, Fable 5 and Fable 5.1) as a coding assistant in developing the software for this study, including the evaluation harness, the training pipeline and the scripts that draw the figures, and to draft and edit parts of the manuscript text. Every figure is drawn by a script from the study's recorded data; no figure was made with an image-generation model. The authors designed the study, reviewed the code and text, checked the reported results against the run records, and take full responsibility for the work. The language models evaluated as baselines and teachers are described in Methods.

\section*{Code and Data availability}
\vspace{-0.7\baselineskip}

The \Prospector{} code is available at \url{https://github.com/lamm-mit/Prospector}.
The semiconductor graph was constructed for SAGE-QA~\citep{hsu2026sageqa}, whose code is available at \url{https://github.com/lamm-mit/SAGE-QA}. The biomaterial graph data is available at \url{https://huggingface.co/lamm-mit/bio-graph-1K/blob/main/large_graph_simple_giant.graphml}. The PFAS and materials-properties graphs were originally released with GraphAgents~\citep{stewart2026graphagents} at \url{https://github.com/lamm-mit/GraphAgents}.

\section*{Competing Interests}
\vspace{-0.7\baselineskip}
The authors declare that they have no competing interests.

\section*{Funding Acknowledgment}
This work is supported by Taiwan Semiconductor Manufacturing Company (TSMC). Additional support is provided by MIT's Generative AI Impact Consortium (MGAIC). We also acknowledge partial compute support from MIT Office of Research Computing and Data (ORCD) and a Thinking Machines Lab research grant. We thank Yu-Chuan Hsu for helpful discussions.

\bibliographystyle{unsrtnat}
\bibliography{refs}

\clearpage
\appendix
\setcounter{table}{0}
\renewcommand{\thetable}{A\arabic{table}}
\setcounter{figure}{0}
\renewcommand{\thefigure}{A\arabic{figure}}

\section{System prompt}
\label{app:prompt}
The policy is served under the system prompt below, reproduced from one evaluation episode on the semiconductor graph. The line giving the number of relation names is generated from the graph the policy is reading, so each of the four graphs announces its own count.

\begin{Verbatim}[fontsize=\scriptsize,baselinestretch=0.95,frame=single,framesep=6pt]
You are a graph explorer. You answer questions by exploring a knowledge graph with tools. You
  cannot see the graph — you only see what the tools return.

TOOLS
find(name)                       -> up to 10 nodes whose name matches `name`, with
                                    type and connection count. Returns no edges. Use it
                                    first to get the exact spelling of an entity.
expand(node, relation?, offset?) -> the edges connected to `node` in either direction,
                                    at most 40 per call. `relation` filters, `offset`
                                    pages. A node with more than 150 connections returns
                                    a relation summary instead of edges — narrow first.
connect(a, b, max_hops?)         -> up to 3 shortest paths between two nodes, as
                                    edges. Use it when you already know both ends.
This graph has 5,289 distinct relation names — too many to list, and they are free text rather
  than a fixed schema. Read the exact relation spelling off the OBSERVATION lines before using it
  in a filter or a citation.

PROTOCOL
On every turn, either call ONE tool or give your final answer.

To call a tool, end your message with exactly one block:
<tool_call>{"tool": "<name>", "args": {...}}</tool_call>
You may reason first inside <think>...</think>. Keep it brief, and write nothing after the tool
  call — the result comes back to you as an OBSERVATION message.

When you have enough evidence, end with your final answer instead:
<think>...</think>
<graph_json>{"nodes": ["..."], "edges": [{"source": "...", "relation": "...", "target":
  "..."}]}</graph_json>
ANSWER: <concise answer>

RULES
1. Every edge in <graph_json> must be an edge you actually saw in an OBSERVATION, copied verbatim
  — same spelling, same relation, same direction.
2. Never invent an edge, a node name, or a relation. If you did not see it, it does not exist for
  you.
3. Cite only the edges that support your answer: the chain from the entity in the question to the
  answer.
4. Tool calls are limited. Prefer the fewest expansions that reach the answer.
5. If you run out of budget, answer with what you have and say what is missing.
\end{Verbatim}

\section{Example episode}
\label{app:episodes}
The episode of Figure~\ref{fig:overview}, a sampled episode of \Prospector{}-4B on the semiconductor graph, is reproduced below from the question to the chain it cites. \texttt{USER}, \texttt{ASST} and \texttt{OBS} mark the question, the policy's own turns and the tool observations, and the line closing each observation counts the edges new to that call, the edges seen so far and the calls remaining.


\begin{Verbatim}[fontsize=\scriptsize,baselinestretch=0.95,frame=single,framesep=6pt]
item medium-0002 · gold path 3 edges · answer set ["wafer-holding system"] · run poolA-s011

USER  Question: In gas-assisted focused electron beam and ion beam processing used in semiconductor
        manufacturing, identify the system responsible for transferring particles to the region of the wafer
        involved in that processing.

      Find the entity named in the question, then explore outward.
      Budget: 20 tool calls.

ASST  <think>Let's search for the entity "gas-assisted focused electron beam and ion beam processing" using the
        `find` tool.</think>
      <tool_call>{"tool": "find", "args": {"name": "gas-assisted focused electron beam and ion beam
        processing"}}</tool_call>

OBS   (call 1/20, find):
      FOUND 6 node(s) matching "gas-assisted focused electron beam and ion beam processing":
        gas-assisted focused electron beam and ion beam processing  [method, 3 connections]
        electron beam  [energy source, 5 connections]
        electron  [particle, 2 connections]
        ion beam  [system, 6 connections]
        gas  [material, 14 connections]
        Ion  [concept, 2 connections]
      No edges returned — use expand() to see connections.
      [0 new edges; 0 edges seen; 19 tool calls left]

ASST  <think>Let's expand `gas-assisted focused electron beam and ion beam processing` to see its
        connections.</think>
      <tool_call>{"tool": "expand", "args": {"node": "gas-assisted focused electron beam and ion beam
        processing"}}</tool_call>

OBS   (call 2/20, expand):
      Cunge, G., Inglebert, R.L., Joubert, O., and Vallier, L. (2002) --[reviewed]--> gas-assisted focused
        electron beam and ion beam processing
      gas-assisted focused electron beam and ion beam processing --[published in]--> J. Vac. Sci. Technol. A
      gas-assisted focused electron beam and ion beam processing --[used for]--> fabrication
      [3 new edges; 3 edges seen; 18 tool calls left]

ASST  <think>Let's expand `fabrication` or check related terms. Let's find nodes related to wafer, particle
        transfer, or similar processing in semiconductor manufacturing.</think>
      <tool_call>{"tool": "find", "args": {"name": "transfer"}}</tool_call>

OBS   (call 3/20, find):
      FOUND 10 node(s) matching "transfer":
        Transfer curves  [data, 2 connections]
        Pattern transfer  [process, 2 connections]
        transfer chamber  [equipment component, 10 connections]
        EUV pattern transfer  [process, 6 connections]
        pattern transfer etch  [domain, 2 connections]
        energy transfer equation  [physical property, 8 connections]
        n-type transfer behaviors  [material, 4 connections]
        sidewall image transfer (SIT)  [term, 5 connections]
        La-Vg transfer characteristics  [characteristic, 3 connections]
        Id –Vg transfer characteristics  [technical_term, 11 connections]
      No edges returned — use expand() to see connections.
      [0 new edges; 3 edges seen; 17 tool calls left]

ASST  <think>Let's expand `transfer chamber` to see what it does and if it's related to wafer or semiconductor
        manufacturing.</think>
      <tool_call>{"tool": "expand", "args": {"node": "transfer chamber"}}</tool_call>

OBS   (call 4/20, expand):
      transfer chamber --[contains]--> isolation trenches
      transfer chamber --[contains]--> wafer-holding system
      transfer chamber --[used between carriers]--> carrier cleaning
      transfer chamber --[maintains]--> ultrahigh vacuum
      transfer chamber --[contains]--> transport system
      transfer chamber --[contains]--> gate valve
      transfer chamber --[contains]--> inner reactor chamber wall
      transfer chamber --[affects]--> aluminum reflow
      multi-chamber dry etching system --[contains]--> transfer chamber
      transfer chamber --[location of]--> particle generation
      [10 new edges; 13 edges seen; 16 tool calls left]

ASST  <think>Let's expand `transport system` and `wafer-holding system` to see their connections.</think>
      <tool_call>{"tool": "expand", "args": {"node": "transport system"}}</tool_call>

OBS   (call 5/20, expand):
      wafer-holding system --[increases]--> transport system
      standard deviation --[enables]--> transport system
      transport system --[affected by]--> particle generation
      wafer --[is transported by]--> transport system
      HCI --[adsorbs on]--> transport system
      transport system --[must suppress]--> electrostatic chuck
      transfer chamber --[contains]--> transport system
      transport system --[affects]--> wafer
      transport system --[generates]--> particles
      mechanical transport with wafer-handling robots --[improves]--> transport system
      [9 new edges; 22 edges seen; 15 tool calls left]

ASST  <think>Let's also expand `wafer-holding system` to see what it connects to.</think>
      <tool_call>{"tool": "expand", "args": {"node": "wafer-holding system"}}</tool_call>

OBS   (call 6/20, expand):
      drying --[subset_of]--> wafer-holding system
      wafer-holding system --[attracts]--> particles
      Table 2.3 --[describes_problems_of]--> wafer-holding system
      wafer-holding system --[optimizes]--> charge build-up
      wafer-holding system --[transfers particles to]--> wafer back side
      conveyor belt --[was replaced by]--> wafer-holding system
      wafer-holding system --[related_to]--> water
      wafer-holding system --[transfers_particles_to]--> wafer
      personnel --[isolated from]--> wafer-holding system
      wafer --[is held by]--> wafer-holding system
      wafer-holding system --[utilizes]--> cluster tools
      wafer-holding system --[removes]--> killer particle diameter
      wafer-holding system --[causes]--> metallic contamination
      transfer chamber --[contains]--> wafer-holding system
      wafer-holding system --[contributes to]--> particle generation
      surface charge analysis equipment --[enables development of]--> wafer-holding system
      wafer-holding system --[utilizes]--> through-the-wall tools
      materials used in the wafer-holding system --[part_of]--> wafer-holding system
      wafer-holding system --[increases]--> transport system
      wafer-holding system --[used in]--> photolithography
      [18 new edges; 40 edges seen; 14 tool calls left]

ASST  <think>Let's check the relation:
      `wafer-holding system --[transfers particles to]--> wafer back side`
      `wafer-holding system --[transfers_particles_to]--> wafer`
      `transfer chamber --[contains]--> wafer-holding system`

      The question asks:
      "In gas-assisted focused electron beam and ion beam processing used in semiconductor manufacturing,
        identify the system responsible for transferring particles to the region of the wafer involved in that
        processing."

      The system is the `wafer-holding system` (which transfers particles to the wafer / wafer back side, and is
        contained in the transfer chamber).

      Let's do a quick connect between `gas-assisted focused electron beam and ion beam processing` and
        `wafer-holding system` or `transfer chamber` just to see if there's any direct path.</think>
      <tool_call>{"tool": "connect", "args": {"a": "gas-assisted focused electron beam and ion beam processing",
        "b": "wafer-holding system"}}</tool_call>

OBS   (call 7/20, connect):
      PATH 1 (3 hops):
      gas-assisted focused electron beam and ion beam processing --[used for]--> fabrication
      wafer back side --[is_part_of]--> fabrication
      wafer-holding system --[transfers particles to]--> wafer back side
      [1 new edges; 41 edges seen; 13 tool calls left]

ASST  <think>The path is:
      gas-assisted focused electron beam and ion beam processing --[used for]--> fabrication
      wafer back side --[is_part_of]--> fabrication
      wafer-holding system --[transfers particles to]--> wafer back side

      Everything is clear and verified directly.</think>
      <graph_json>{
        "nodes": [
          "gas-assisted focused electron beam and ion beam processing",
          "fabrication",
          "wafer back side",
          "wafer-holding system"
        ],
        "edges": [
          {
            "source": "gas-assisted focused electron beam and ion beam processing",
            "relation": "used for",
            "target": "fabrication"
          },
          {
            "source": "wafer back side",
            "relation": "is_part_of",
            "target": "fabrication"
          },
          {
            "source": "wafer-holding system",
            "relation": "transfers particles to",
            "target": "wafer back side"
          }
        ]
      }</graph_json>
      ANSWER: wafer-holding system
\end{Verbatim}

\end{document}